\documentclass[conference]{IEEEtran}
\IEEEoverridecommandlockouts
\usepackage{cite}
\usepackage{multirow}
\usepackage{amsmath,amssymb,amsfonts}
\usepackage{algorithmic}
\usepackage{graphicx}
\usepackage{textcomp}
\usepackage{url}
\usepackage{xcolor}
\usepackage{comment}
\usepackage{tabularx}
\def\BibTeX{{\rm B\kern-.05em{\sc i\kern-.025em b}\kern-.08em
    T\kern-.1667em\lower.7ex\hbox{E}\kern-.125emX}}
\begin{document}

\title{Preliminary Systematic Literature Review of Machine Learning System Development Process}

\author{
\IEEEauthorblockN{Yasuhiro Watanabe$^{1}$, Hironori Washizaki$^{2}$, Kazunori Sakamoto$^{3}$, \\ Daisuke Saito$^{4}$, Kiyoshi Honda$^{5}$, Naohiko Tsuda$^{6}$, Yoshiaki Fukazawa$^{7}$ and Nobukazu Yoshioka$^{8}$}
\IEEEauthorblockA{
\textit{$^{1, 2, 3, 4, 6, 7}$Waseda University}, Tokyo, Japan \\
\textit{$^{5}$Osaka Institute of Technology}, Osaka, Japan \\
\textit{$^{8}$National Institute of Informatics}, Tokyo, Japan \\
$\{^{1}$jellyfish44-time@akane., $^{2}$washizaki@, $^{3}$k.sakamoto@aoni., $^{4}$d.saito@fuji., $^{6}$821821@toki., $^{7}$fukazawa@$\}$ waseda.jp,\\ $^{5}$kiyoshi.honda@oit.ac.jp,
$^{8}$nobukazu@nii.ac.jp}
}

\maketitle

\begin{abstract}
Previous machine learning (ML) system development research suggests that emerging software quality attributes are a concern due to the probabilistic behavior of ML systems.
Assuming that detailed development processes depend on individual developers and are not discussed in detail.
To help developers to standardize their ML system development processes, we conduct a preliminary systematic literature review on ML system development processes.
A search query of 2358 papers identified 7 papers as well as two other papers determined in an ad-hoc review.
Our findings include emphasized phases in ML system developments, frequently described practices and tailored traditional software development practices.

\end{abstract}


\section{Introduction}
\label{intro}
Machine learning (ML) systems are complex systems.
Machine learning algorithm's behaviors are probabilistic because they depend on training data.
In contrast, the behaviors in traditional software development are defined by code.

Some researchers suspect that the probabilistic behaviors derive emerging quality or validation concerns \cite{Arpteg,Antoniol,Lipton2018}.
Additionally, other studies have investigated ML system's bugs derived from algorithms, data dependency and its architectures~\cite{HiddenTechDebt,Thung}.
Moreover, data infrastructure is also complex because it often deals with big data \cite{ZHAO}.
Quality aspects of ML systems are well documented.
A literature review is conducted~\cite{Hosokawa}.
ML system development challenges are also discussed by conducting empirical case study~\cite{Lwakatare}.
However, development practices are not discussed well.
We assume that these practices depend on individual developers or organizations.
It prevents developers from utilizing practices to ML system developments.

To understand current practices and to help developers standardize ML system development processes, we implement a preliminary systematic literature review.
We defined a search query in Scopus for the review. The query returned 2358 papers as potential hits.
Of those, 7 papers are related to ML system development process.
We reviewed and compared these seven papers to two other papers which were retrieved by an ad-hoc review.
This paper reveals seven main findings:
\begin{itemize}
\item F1: Model Training \& Evaluation phases are frequently described in several papers,
\item F2: Some phases are mentioned in specific communities,
\item F3: Some issues and practices are described frequently,
\item F4: ML system development employs practices in Data Concerns, Start Small or Measure category in several phases,
\item F5: Practices in Separation of Concerns category are employed even in ML system development,
\item F6: Practices in Goal-oriented category are tailored to Model Evaluation and
\item F7: Traditional practices are useful even in ML system developments.
\end{itemize}

\section{Review Method}
\label{method}

\subsection{Research Questions}
To help developers standardize ML system development processes,
We defined the three research questions (RQs).

To understand how existing practices tackle issues in ML system developments, we investigated:
\textbf{RQ1: Which phases are emphasized in ML system developments?}

To understand detailed activities, we investigated:
\textbf{RQ2: What kind of practices are included in ML system developments?}

To help developers to utilize their current experiences even in ML system developments, we investigated:
\textbf{RQ3: Do practices in ML system developments include traditional software development practices?}

\subsection{Inclusion and Exclusion Criteria}
We surveyed following articles which satisfy all following inclusion criteria:
\begin{itemize}
  \item Indexed in Scopus,
  \item Related to Software engineering process for Machine Learning system,
  \item Related to processes, practices or role of developers in ML system development,
  \item Included in Engineering, Computer Science or Mathematics domain, and
  \item Published in January 2010 to March 2019
\end{itemize}

We excluded articles which satisfy any following exclusion criteria:
\begin{itemize}
  \item Proposals of machine learning algorithm
  \item Machine learning applications for software engineering tasks or a specific project
  \item Proposals of software engineering techniques without tailoring a process or phases (e.g., proposal of tool to improve efficiency of a specific phase)
  \item Quality concerns of ML systems
\end{itemize}

\subsection{Search Strategies}
We constructed the following query in Scopus to satisfy the above criteria.
Our query in Scopus was \textit{((``machine learning''  OR  ``machine-learning''  OR  ``ML'' )  AND  (``software''  OR  ``system''  OR  ``systems'' )  AND ``engineering'') AND (``practice''  OR  ``organization pattern'' OR ``process'' OR ``process pattern'' OR ``empirical'' OR ``empirical study''  OR  ``case study''  OR  ``field study'')}.
Its search targeted the title and abstract.

The query returned 2358 papers from Scopus.
The initial results included papers related to machine learning algorithms, applications to a specific domain, machine learning utilization to software engineering, etc.
A review by first author identified seven papers related to ML system development processes.
Additionally, the first author found two related papers, which are not indexed in Scopus via a ad-hoc review.
Thus the first author analyzed the nine papers and identified issues and practices contained within.

\section{Results and Findings}
\label{result}

\subsection{Overall findings}
We surveyed nine papers (\cite{S1, S2, S3, S4, S5, S6} via Scopus and \cite{S7, A1, A2} via ad-hoc review).
The surveyed papers include five papers based on interview with developers (~\cite{S1,S2,S3,S5,A1}~) and three papers proposed their own processes (~\cite{S4, S6, S7}~).
Other one paper provides the best practices from industry (~\cite{A2}~).
High distribution of interview papers implies that current research tried to figure out existing development processes in bottom-up by conducting interviews.

Five papers are published in computer science communities (~\cite{S1,S2,S5,A1,A2}~) and four papers are published in human-centered design communities (~\cite{S3, S4, S6, S7}~).

Reviewed studies assume that ML system development teams include the following roles: developers (~\cite{A1,A2,S1}~), data scientists (~\cite{S2,S5}~), UX designers (~\cite{S3}~), domain experts (~\cite{S6}~) and end-users (~\cite{S4,S7}~).

To answer RQ1, RQ2 and RQ3,
We mapped issues and practices described in each papers into phases of \cite{A1} (Table \ref{tab:issue_practice}).
Phases and Definitions in Table \ref{tab:issue_practice} show phases and their definitions in ML system development processes.
Issues and Practices in Table \ref{tab:issue_practice} show issues and practices which we found in each phase by reviewing nine papers.

To answer RQ2 and RQ3, we defined categories of practices, their definitions and distribution of categories by phase (Table~\ref{tab:category}).
We found that several practices have the same purpose or characteristics even though these practices are employed in different phases.
We defined categories based on some common purposes or characteristics in practices (Categories and Definitions in Table~\ref{tab:category}) and categorized practices into a category by phase (Practices by Phase in Table~\ref{tab:category}).

\begin{table*}[h]

\caption{ML system development issues and practices by phase}
\begin{center}
\begin{tabular}{|p{2.1cm}|p{2cm}|p{6.4cm}|p{6.5cm}|}
\hline
\textbf{Phase \cite{A1}} & \textbf{Definitions}& \textbf{\textit{Issues}} & \textbf{\textit{Practices}}\\ \hline
  \begin{tabular}{l}
    Model \\ Requirement \\ (MR)
  \end{tabular} &
  \begin{tabular}{l}
    Decide the \\ functions and \\ types of models
  \end{tabular} &
  \begin{tabular}{l}
    I11: For designers, difficult to understand \\ML algorithm and its potential \cite{S3} \\
    I12: Lack of prototyping methodology~\cite{S3} \\
    I13: End-users' ignorance of ML algorithms~\cite{S7}\\
    I14: End-users' indirect interaction with ML \cite{S7}\\
    I15: Lack of Ethical Considerations \cite{S3}
  \end{tabular} &
  \begin{tabular}{l}
    P11: First, design and implement metric \cite{A1,A2}\\
    P12: Choose a simple, observable and attributable \\ metric for a goal \cite{A2}\\
    P13: Gather the users' voice \cite{A2}\\
    P14: Turn heuristics into features (understand task) \cite{A2}\\
    P15: Apply ML in less obvious ways to envision \\ opportunity \cite{S3} \\
    P16: Represent ML’s Dependency on Data in Early \\ Prototypes (Estimate the difficulties in Data collection \\and labeling) \cite{S3}\\
    P17: Workshops to elicit domain knowledge \cite{S6}

  \end{tabular} \\ \hline

  \begin{tabular}{l}
    Data Collection \\ (DCo)
  \end{tabular} &
  \begin{tabular}{l}
    Collect data\\
    or construct\\
    infrastructures \\
    to collect data \\
  \end{tabular} &
  \begin{tabular}{l}
    I21: Difficult to understand data from third party \cite{S1}\\
    I22: Strain on resources to collect data \cite{S1,S3} \\
    I23: Difficult to understand data format \cite{S1,A2}
  \end{tabular}&
  \begin{tabular}{l}
    P21: Launch a product without machine learning \cite{A2}\\
    P22: Give column owners and documentation \cite{A2}\\
    P23: Clean up features you are no longer using \cite{A2}\\
    P24: Tool to map Data and model \cite{A1}
  \end{tabular} \\ \hline

  \begin{tabular}{l}
    Data Cleaning \\ (DCl)
  \end{tabular} &
  \begin{tabular}{l}
    Remove noisy \\ record
  \end{tabular} &
  \begin{tabular}{l}
    I31: The method to preprocess data is not mentioned \\ in a paper \cite{S5}
  \end{tabular} &
  \\ \hline

  \begin{tabular}{l}
    Data Labeling \\ (DL)
  \end{tabular} &
  \begin{tabular}{l}
    Determine labels \\ for each record
  \end{tabular} &
  \begin{tabular}{l}
    I41: Strain on resources for labeling \cite{S1}
  \end{tabular} &
\\ \hline

  \begin{tabular}{l}
    Feature \\ Engineering \\ (FE)
  \end{tabular} &
  \begin{tabular}{l}
    Extract informa- \\
    tive features to \\
    improve model \\
    training
  \end{tabular} &
  \begin{tabular}{l}
    I51: Difficult to measure the effect of feature \cite{A2} \\
    I52: Important features affect the result \cite{A2}
  \end{tabular} &
  \begin{tabular}{l}
    P51: Domain knowledge and past experience \cite{S1} \\
    P52: Observe statistical characteristics of data \cite{S1} \\
    P53: Start with human-understandable features \cite{A2}
  \end{tabular} \\ \hline

  \begin{tabular}{l}
    Model Training \\ (MT)
  \end{tabular} &
  \begin{tabular}{l}
    Train and\\
    tune models
  \end{tabular} &
  \begin{tabular}{l}
    I61: Ad-hoc algorithm selection based on past \\ experience \cite{S1}\\
    I62: Ignorance of new technologies that do not \\require data \cite{S2}\\
    I63: No significant improvement by using current \\features \cite{A2}\\
    I64: Difficult to reuse the model for other domains \\or data formats \cite{A1}\\
    I65: Complex component entanglement \cite{S1}
  \end{tabular} &
  \begin{tabular}{l}
    P61: Select simplest algorithm \cite{S1,A2,S5} \\
    P62: Separate functions into different models \cite{A2}\\
    P63: Find new features to improve models \cite{A2} \\
  \end{tabular} \\ \hline

  \begin{tabular}{l}
    Model Evaluation \\ (ME)
  \end{tabular} &
  \begin{tabular}{l}
    Verify models\\
    based on\\
    metrics
  \end{tabular} &
  \begin{tabular}{l}
    I71: Difficult to understand the results ~\cite{S1,S2,S6}\\
    I72: Training and serving skewed \\(unpredictable behavior) \cite{A2,S3}\\
    I73: Frequent revisions initiated by model changes, \\parameter tuning, and data updates ~\cite{A1}\\
    I74: Identical short-term behaviors do not imply\\ identical long-term behavior ~\cite{A2}\\
  \end{tabular} &
  \begin{tabular}{l}
    P71: Measure the model performance \cite{S2} \\
    P72: Test infrastructure independently \cite{A2}\\
    P73: Measure the delta between the current model \\and a baseline \cite{S1,A2}\\
    P74: Focus on the ultimate objective metrics \cite{A2}\\
    P75 Evaluate models based on users' own criteria \cite{S4}
  \end{tabular} \\ \hline

  \begin{tabular}{l}
    Model \\ Deployment (MD)
  \end{tabular} &
  \begin{tabular}{l}
    Deploy models\\
    to devices
  \end{tabular} &
  \begin{tabular}{l}
    I81: Copy pipeline and drop necessary data \cite{A2}\\
    I82: Various launch criteria \cite{A2}\\
    I83: Concerned with other modules \cite{A1}
  \end{tabular} & \begin{tabular}{l}
    P81: Model the infrastructure correctly \cite{A1,A2}\\
    P82: Evaluate models before releases \cite{A2}\\
    P83: Keep on single criterion \cite{A2}
  \end{tabular} \\ \hline

  \begin{tabular}{l}
    Model \\ Monitoring (MM)
  \end{tabular} &
  \begin{tabular}{l}
    Monitor models'\\
    behaviour
  \end{tabular} &
  \begin{tabular}{l}
  \end{tabular} &
  \begin{tabular}{l}
    P91: Know the data attribute requirements \cite{A1,A2}\\
    P92: Watch for silent failures \cite{A2}
  \end{tabular} \\ \hline

  \begin{tabular}{l}
    Cross-Cutting \\ (CC)
  \end{tabular} &
  \begin{tabular}{l}
    Concerns in a \\
    whole \\ development
  \end{tabular} &
  \begin{tabular}{l}
    I101: Highly dependent on individuals \cite{S1,S2} \\
    I102: Lack of collaboration between developers \\ and other roles \cite{S3,S6}
  \end{tabular} &
  \begin{tabular}{l}
    P101: Provide the opportunities to share knowledge \\ (e.g, mailing lists, open forums, internal conferences, \\etc.)~\cite{A1,S5}
  \end{tabular} \\ \hline

\end{tabular}
\label{tab:issue_practice}
\end{center}
\end{table*}

\begin{table*}[h]
\caption{Categories of practices}
\vspace{-0.4cm}
\begin{center}
\begin{tabular}{|p{1.5cm}|p{4.0cm}|p{0.7cm}|p{0.7cm}|p{0.7cm}|p{0.7cm}|p{0.7cm}|p{0.7cm}|p{0.7cm}|p{0.7cm}|p{0.7cm}|}
\hline
\textbf{Practice Category} & \textbf{Category Definition} &\multicolumn{9}{|c|}{\textbf{Practices by Phase (each phase is represented by its abbreviation)}} \\ \cline{3-11}
 &  & MR & DCo & DCl & DL & FE & MT & ME & MD & MM\\ \hline \hline
Start Small & Practices to start with simplified issues. &  & P21 & & & P53 & P61 & & P83 &  \\ \hline
Traditional Practice & Practices employed even in traditional software developments & P13, P14 & P22, P23 & & & & & & P82 & \\ \hline
Goal-oriented & Practices to focus on a goal of the project & P11 & & & & & & P74 & & \\ \hline
Data Concerns & Practices to deal with issues related to data & P16 & P24 & & & P52 & P63 & & & P91 \\ \hline
Separation of Concerns & Practices to identify sub issues and deal with them step-by-step & & & & & & P62 & P72 & P81 &  \\ \hline
Measure & Practices to measure uncertainty in developments & P12 & & & & & & P71, P73 & & P92\\ \hline
Heuristic & Practices which rely on developers' experiences & P15 & & & & P51 & & & & \\ \hline
\end{tabular}
\label{tab:category}
\end{center}
\end{table*}


\subsection{RQ1: Phases in ML system developments}
As answer to RQ1, we explain two findings F1 and F2.


\textbf{F1: Model Training \& Evaluation phases are frequently described in several papers.}
Issues in Model Training (MT) phase are described in four papers \cite{S1, S2, A1, A2} (Table~\ref{tab:issue_practice}).
Issues in Model Evaluation (ME) phase are described in six papers \cite{S1, S2, S3, S6, A1, A2} (Table~\ref{tab:issue_practice}).
Practices in MT and ME phases are also described in several papers (Table~\ref{tab:issue_practice}).
Practices in MT are described in three papers \cite{S1,A2,S5} and practices in ME are described in four papers \cite{S1,S2,S4,A2} (Table~\ref{tab:issue_practice}).

\textbf{F2: Some phases are mentioned in specific communities.}
Issues in Model Requirement (MR) phase are mentioned by studies in human-centered design communities.
Five issues in MR phase are described in two papers \cite{S3,S7} which are published only in human-centered design communities (Table~\ref{tab:issue_practice}).

On the other hand, several data and model related phases are described only in computer science communities.
Issues in Data Cleaning (DCl) \& Data Labeling (Dl), practices in Data Collection (DCo), and issues and practices in Feature Engineering (FE), MT, \& Model Deployment (MD) are mentioned by papers which are published only in computer science communities (Table~\ref{tab:issue_practice}).

\subsection{RQ2: Practices in ML system developments}
As answer to RQ2, we explain two findings F3 and F4.

\textbf{F3: Some issues and practices are described frequently.}
Several papers mentioned same issues or practices (Table \ref{tab:issue_practice}).
Issue I71 in MT phase is described in three papers.
Issues I22 \& I23 in DCo phase, I72 in MT phase and I101 \& I102 in Cross-Cutting (CC) phase are described in two papers.
These frequent issues are included in DCo, MT or CC phase.
Issues I101 and I102 are related to human capabilities.
Practice P61 in MT phase is described in three papers.
Practices P11, P73, P81, P91 and P101 are described in two papers.
These frequent practices are included in MR, MT, ME, MD, or Model Monitoring (MM) and CC phases.


\textbf{F4: Practices in Data Concerns, Start Small or Measure category are employed in many phases.}
Practices in Data Concerns category (Table~\ref{tab:category}) are included in five phases (MR, DCo, FE, MT and MM).
Additionally, there is no practice for DCo and DL phases.
We expect that difficulties in data management are tackled but data collection and data labeling are still challenging.

Practices in Start Small or Measure category are also employed in many phases.
Practices in Start Small or Measure category are included in four phases (Table~\ref{tab:category}).

\subsection{RQ3: Traditional software development practices}
As answer to RQ3, we explain following three findings: F5, F6 and F7.
\textbf{F5: Practices in Separation of concerns category are employed even in ML system development.}
Separation of concerns is a traditional concept in software engineering field (e.g. aspect-oriented programming is proposed to help the separation of concerns \cite{kiczales1997aspect}).
Separation of Concerns shows that Separation of Concerns practices are included in MT, ME and MD phases (Table~\ref{tab:category}).
Thus, this concept can be employed even in ML system developments.

\textbf{F6: Practices in Goal-oriented category are tailored to Model Evaluation.}
Goal-oriented practices are often employed to align software developments to business goals~\cite{GQM_S}.
We categorized P11 and P74 into Goal-oriented category (Table~\ref{tab:category}) because these two practices mention goals in ML system developments.
P11 is related to designing metrics for Model Evaluation and P74 is employed in Model Evaluation.
Thus, we argue that practices in Goal-oriented category are tailored to Model Evaluation.

\textbf{F7: Traditional practices are useful even in ML system developments.}
We found other five traditional software engineering practices are described in reviewed papers (Traditional Practice in Table \ref{tab:category}).
These five practices include requirement elicitation from users (P13 and P14), traceability management (P22), refactoring (P23), and testing (P82).
Practices in Traditional Practice category are employed in MR, DCo and MD phases (Table~\ref{tab:category}).

Practices in Separation of concerns, Goal-oriented or Traditional Practice category are employed in five (MR, DCo, MT, ME and MD) out of eight phases (Table~\ref{tab:category}).
We thus argue that traditional concepts or practices may be useful in even though issues in ML system have unique aspects.

\section{Threats to Validity}
\label{threats}
The inclusion and exclusion criteria may be a threat to validity.
In this research, we reviewed papers which purely focused on ML system development processes.
However, case study papers may possibly include discussion about development processes.
Additionally, the boundary between data mining projects and ML system developments is vague, which may affect the authors' paper selection.

\section{Conclusion}
\label{conc}
To understand current practices and to help developers standardize ML system development processes, we conducted a preliminary systematic literature review on ML system development processes.

This study reveals seven following findings: 
F1: Model Training \& Evaluation phases are frequently described in several papers,
F2: Some phases are mentioned in specific communities,
F3: Some issues and practices are described frequently,
F4: ML system development employs practices in Data Concerns, Start Small or Measure category in several phases,
F5: Practices in Separation of Concerns category are employed even in ML system development,
F6: Practices in Goal-oriented category are tailored to Model Evaluation and
F7: Traditional practices are useful even in ML system developments.

In the future, we will conduct interviews with developers and discuss the effectiveness of each practice.

\section{Acknowledgements}
This work was supported by JST-Mirai Program Grant Number JP18077318, Japan.
\bibliographystyle{IEEEtran}
\bibliography{related,survey}

\end{document}